\documentclass{article}
\usepackage{spconf,amsmath,graphicx,makecell,multirow,subfigure,CJKutf8,amsfonts}
\usepackage[T1]{fontenc}
\usepackage[colorlinks,
            linkcolor=black,
            anchorcolor=black,
            citecolor=black]{hyperref}
\usepackage{xcolor}

\title{ANCIENT CHINESE WORD SEGMENTATION AND Part-of-Speech TAGGING USING DISTANT SUPERVISION}
\name{Shuo Feng, Piji Li$^{\ast}$\thanks{*Corresponding author.} }
\address{Nanjing University of Aeronautics and Astronautics, China}
\begin{document}
%
\maketitle
\begin{abstract}
Ancient Chinese word segmentation (WSG) and part-of-speech tagging (POS) are important to study ancient Chinese, but the amount of ancient Chinese WSG and POS tagging data is still rare. In this paper, we propose a novel augmentation method of ancient Chinese WSG and POS tagging data using distant supervision over parallel corpus. However, there are still mislabeled and unlabeled ancient Chinese words inevitably in distant supervision. To address this problem, we take advantage of the memorization effects of deep neural networks and a small amount of annotated data to get a model with much knowledge and a little noise, and then we use this model to relabel the ancient Chinese sentences in parallel corpus. Experiments show that the model trained over the relabeled data outperforms the model trained over the data generated from distant supervision and the annotated data. Our code is available at https://github.com/farlit/ACDS.
\end{abstract}
\section{Introduction}
\label{sec:intro}

Word segmentation (WSG) and part-of-speech (POS) tagging are two fundamental lexical analysis problems in ancient Chinese natural language processing (NLP) \cite{li2022first}. Automatic WSG and POS tagging have achieved great progress using supervised learning on large labeled corpus. However, the rareness of labeled ancient Chinese corpus and the huge difficulty of manual annotation on it limit further improvement of model performance on the joint WSG and POS tagging task. 

\begin{CJK*}{UTF8}{bsmi}

In EvaHan2022\footnote{https://circse.github.io/LT4HALA/2022/EvaHan\label{1}}, participants have proposed useful ways of data augmentation. One way is to use autoregressive language model to generate pseudo-data based on linearized labeled sentences \cite{ding2020daga}\cite{shen2022data}, where linearizing the sentences is to insert the tags before the corresponding Chinese characters. Another way is to ask pretrained model to predict masked words to mine ancient Chinese knowledge \cite{shen2022data}. The last way is to randomly mask the input words during training \cite{tian2022ancient}.
\begin{figure}[t]
  \vspace{0.45cm}
  \centering
  \centerline{\includegraphics[width=8.8cm]{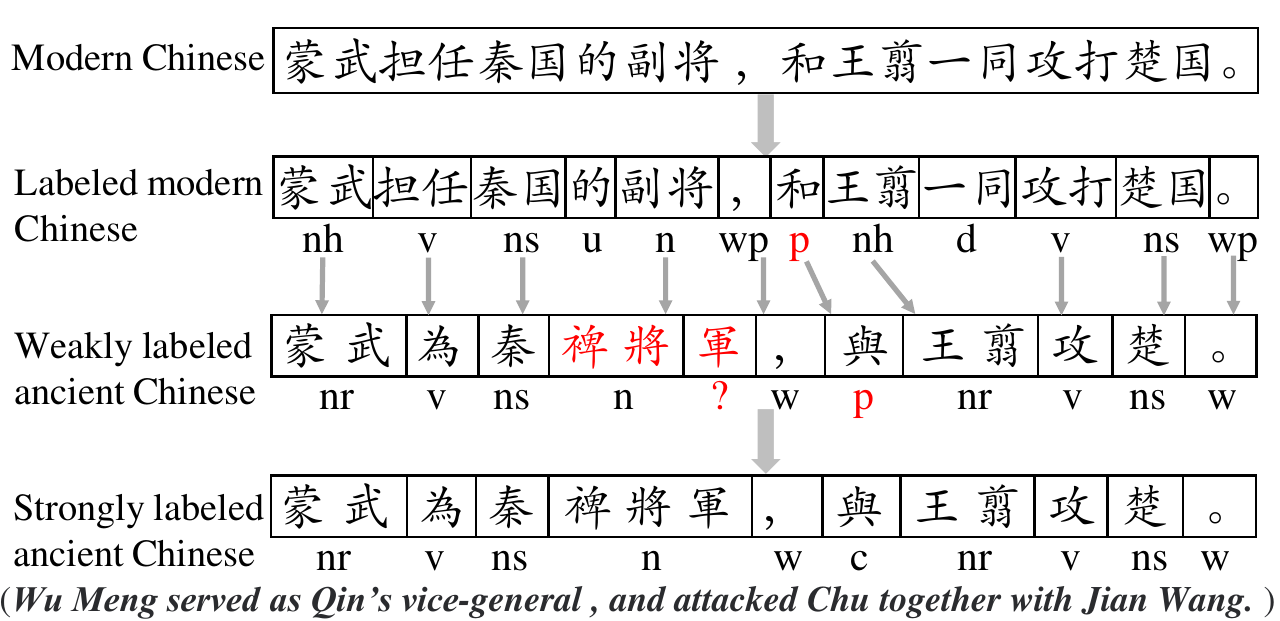}}
  \vspace{-0.2cm}
%
\caption{
An example of automatic annotation of ancient Chinese using distant supervision. `?' represents Chinese words not tagged. Red Chinese words indicate wrong word segmentation. Red English characters indicate wrong POS tagging results.}
\vspace{-0.3cm}
\label{fig:inro-example}
\end{figure}
However, the above approaches can only generate samples similar to the given annotated data or implicitly augment data, which can not solve the problem of data scarcity essentially, and their implementation processes are complicated.

In low-resource scenario, distant supervision can provide ample but inaccurately labeled samples at low cost \cite{mintz2009distant}. A widely used method of distant supervision is to utilize parallel corpus for word alignment. Although annotated data of low-resource languages is difficult to obtain, parallel corpora of low-resource languages and high-resource languages are plentiful. Word alignment can project annotations from the high-resource language to the low-resource language, which has been proved effective in WSG \cite{che2019word}, POS tagging \cite{yarowsky2001inducing}\cite{fang2016learning}, named entity recognition (NER) \cite{wang2014cross}, and dependency parsing \cite{mcdonald2013universal}. Although annotation projection is effective in data augmentation, it has several problems including errors and omissions in tagging the low-resource languages \cite{das2011unsupervised}\cite{tackstrom2013token}. As shown in Figure \ref{fig:inro-example}, the word ``裨將軍'' (vice-general) is wrongly segmented as two words ``裨將'' (vice-general) and ``軍'' (army); the word ``與'' (with), whose POS tag in this sentence is `c' (conjunction), is wrongly tagged as `p' (preposition); the word ``軍'' (army) is not tagged. Inspired by the memorization effects of deep neural networks (DNNs), which tend to fit clean data firstly and then fit noisy data gradually \cite{arpit2017closer}\cite{bai2021understanding}, we use the large data obtained from word alignment and the small annotated data to get a model with much knowledge and a little noise. Then we use the model to relabel the large data to reduce its errors and omissions. 

As shown in Figure \ref{fig:inro-example}, we propose a novel data augmentation of ancient Chinese WSG and POS tagging completed in three steps. First, we use Chinese NLP tool to perform WSG and POS tagging on modern Chinese sentences to get sentences with word boundaries and POS tags. Then, we project the word boundaries and POS tags from modern Chinese to ancient Chinese. After that, we train the SIKU-Roberta\footnote{https://huggingface.co/SIKU-BERT/sikuroberta\label{2}} over the large weakly labeled WSG and POS tagging data obtained from distant supervision to get the first stage model. We continue to train the first stage model over the small manually annotated WSG and POS tagging data to get the second stage model. At last, we use the second stage model to relabel the large weakly labeled data generated from distant supervision.

Our work solves the problem of data scarcity in ancient Chinese WSG and POS tagging. We make four main contributions: (1) We introduce distant supervision in ancient Chinese WSG and POS tagging. (2) We use a parallel corpus to generate large ancient Chinese WSG and POS tagging data using distant supervision. (3) We propose a novel method of denoising and completing the labels of the large weakly labeled data generated from distant supervision by relabeling it. (4) Extensive experiments demonstrate the effectiveness of the method of distant supervision and relabeling.
\section{Method}
\label{sec:format3}

\subsection{Overview}
\label{ssec:overview}

For ancient Chinese WSG and POS tagging, we use a POS tagging set with 22 tags\footnote{https://github.com/CIRCSE/LT4HALA/blob/master/2022\ /data\_and\_doc/EvaHan\_guidelines\_v1\_training.pdf\label{3}}, including verb(v), noun(n), adjective(a), person(nr), etc., and a WSG tag set \{B, M, E, S\}, where `B', `M', `E' represent the beginning, middle and end of a word respectively, and `S' indicates a word only with one character. Compared with pipeline method, joint implementation of ancient Chinese WSG and POS can improve the performance of the two tasks \cite{shi2010crf}. Hence, we treat the ancient Chinese WSG and POS tagging as one sequence labeling task using a hybrid tag set, which contains 88 tags.

\begin{figure}[t]
  \vspace{0.05cm}
  \centering
  \centerline{\includegraphics[width=8.8cm]{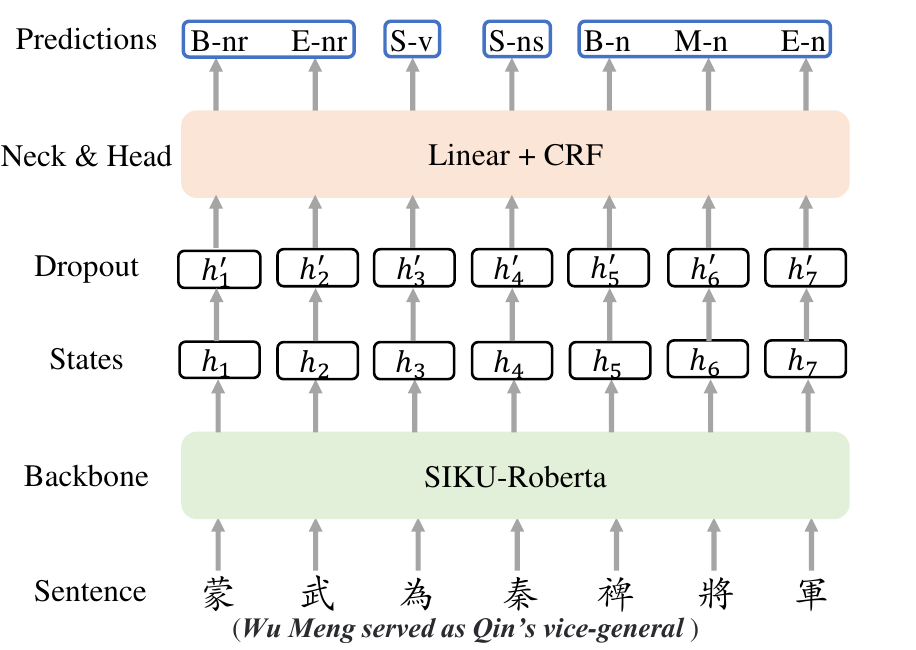}}
  \vspace{-0.25cm}
%
\caption{
 The overall structure of our model.}
\vspace{-0.3cm}
\label{fig:model}
\end{figure}

Our model, which is shown in Figure \ref{fig:model}, consists of a backbone, SIKU-Roberta\textsuperscript{\ref{2}}, one linear layer and one Conditional Random Fields (CRF) \cite{lafferty2001conditional} layer for the joint WSG and POS tagging sequence labeling. Considering an ancient Chinese sentence $X$ containing $n$ characters, formally denoted as \{$x_1$,$x_2$,...,$x_{n}$\}, where $n$ is the length of $X$, our task is to use our model to get its corresponding hybrid tag sequence $Y$, denoted as \{$y_1$,$y_2$,...,$y_{n}$\}. For example, in Figure \ref{fig:model}, the input sentence is ``蒙武為秦裨將軍'' (Wu Meng served as Qin’s vice-general), and its corresponding ouput sequence of hybrid tags is ``B-nr E-nr S-v S-ns B-n M-n E-n''.

\subsection{Sequence Labeling via CRF}
\label{ssec:subhead0}

Given the sentence $X$, SIKU-Roberta\textsuperscript{\ref{2}} can produce its hidden states $\mathbf{H}\in\mathbb{R}^{n\times d}$, where d is the hidden layer size of SIKU-Roberta\textsuperscript{\ref{2}}. The hidden state $\mathbf{h}_{i}\in\mathbb{R}^{d}$ is fed into the linear layer to compute the predicted logit vector $\mathbf{s}_{i}\in\mathbb{R}^{v}$ for the CRF \cite{lafferty2001conditional} layer. The emission scores are computed as follows,
\begin{equation}
    \mathbf{s}_{i} = \mathbf{h}_{i}^{\mathbb{T}}\mathbf{W}_{s}+\mathbf{b}_{s}
\end{equation}
where $\mathbf{W}_{s}\in\mathbb{R}^{d\times v}$, $\mathbf{b}_{s}\in\mathbb{R}^{v}$ are the weight matrix and the bias of the linear layer, and ${v}$ is the number of hybrid tags. And then the likelihood of the target sequence $Y$ via CRF \cite{lafferty2001conditional} is constructed as:
\begin{equation}
    P_{crf}(Y|X)=\frac{1}{Z(X)}\mathbf{exp}\bigg(\sum\limits_{i=1}^n s(y_{i})+\sum\limits_{i=2}^n t(y_{t-1},y_{t})\bigg)
\end{equation}
where $Z(X)$ is the normalizing factor and $s(y_{i})$ respents the label score of $y$ at position $i$, which can be obtained from the predicted logit vector $\mathbf{s}_{i}\in\mathbb{R}^{v}$. The value $t(y_{t-1},y_{t})=\mathbf{M}_{y_{t-1},y_{t}}$ denotes the transition score from token $y_{t-1}$ to $y_{t}$ where $\mathbf{M}\in\mathbb{R}^{{v}\times{v}}$ is the transition matrix, which can be learnt as
neural network parameters during the end-to-end
training procedure. The loss function $L$ is denoted as:
\begin{equation}
    L=-\mathbf{log}P_{crf}(Y|X)
\label{fun:loss}
\end{equation}
We update parameters of our model by minimize the loss function Equation (\ref{fun:loss}). In the inference stage, for the input sentence $X$, we use the Viterbi algorithm \cite{lafferty2001conditional}\cite{forney1973viterbi} to obtain the predicted optimal hybrid tag sequence.

\subsection{Distant Supervision via Word Alignment}
\label{ssec:subhead1}

Although vast annotated ancient Chinese data is difficult to obtain, the parallel corpus of modern Chinese and ancient Chinese is sufficient, which provides the opportunity for obtaining ample ancient Chinese labeled data using distant supervision by alignment processing. The modern Chinese in parallel corpus are unlabeled, so we first use LTP \cite{che2010ltp} to perform WSG and POS tagging on the modern Chinese to get sentences with word boundaries and POS tags, and then divide the ancient Chinese sentences into single characters. 
\begin{figure}[t]
  \vspace{0.05cm}
  \centering
  \centerline{\includegraphics[width=8.8cm]{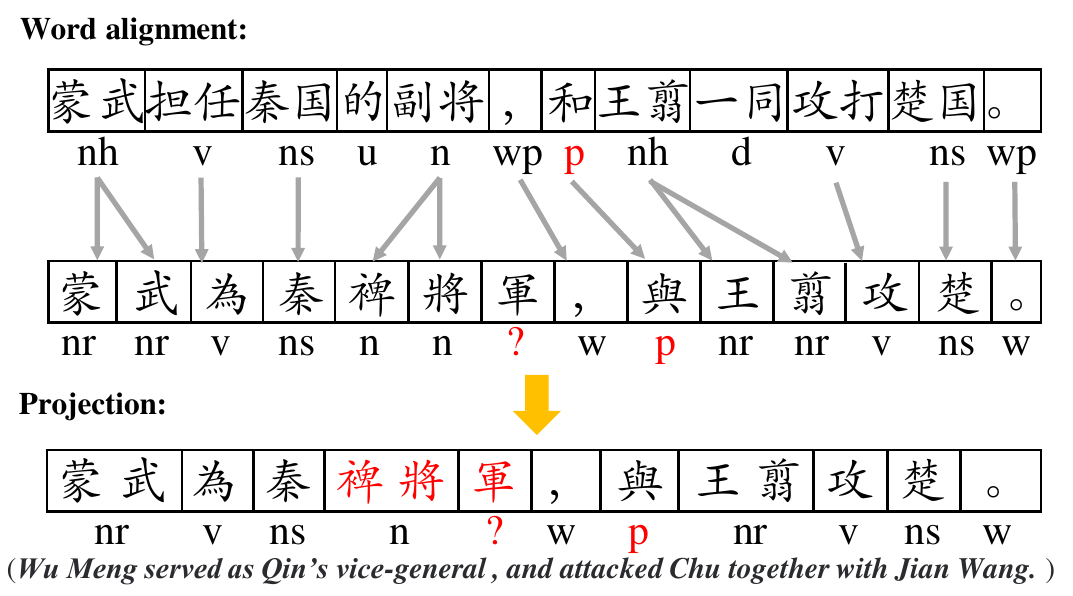}}
  \vspace{-0.3cm}
%
\caption{
 An example of word alignment.}
\vspace{-2px}
\label{fig-wa}
\end{figure}

After processing the parallel corpus, we set modern Chinese as source language and ancient Chinese as target language. Then we use GIZA++ to implement word alignment on the parallel corpus, using IBM model 4 \cite{della1994mathematics}, an unsupervised generative model, which can find possible pairs of aligned words and calculate their alignment probabilities.
\begin{table}[htbp]
\centering
\scalebox{1.18}
{
	\begin{tabular}
	{|c|c|c|c|c|c|}\hline
	a:a & b:a & c:c & d:d & e:y & h:null\\ \hline

    i:null & j:null & k:null & m:m & n:n & nd:f\\ \hline
	
    nh:nr & ni:ns & nl:n & ns:ns & nt:t & nz:n\\ \hline

	o:s & p:p & q:q & r:r & u:u & v:v\\ \hline
	
	wp:w & ws:x & x:null &  g:null & z:a & -\\ \hline
	\end{tabular}
}
\caption{The dictionary of modern Chinese POS tags and ancient Chinese POS tags. The keys are POS tags of modern Chinese. The values are POS tags of ancient Chinese.}
\vspace{-2px}
\label{tab:dic}
\end{table} 

LTP \cite{che2010ltp} adopts the 863 POS tagging set \cite{jinteng2013construction}, containing 29 types of POS tags. The types of ancient Chinese POS tags\footnotemark[3] is 22. As shown in Table \ref{tab:dic}, we set up a POS tagging dictionary, where the keys are the POS tags adopted by LTP \cite{che2010ltp} and the values are the POS tags of ancient Chinese. After implementing the IBM model 4 \cite{della1994mathematics}, we obtain possible aligned pairs with alignment probabilities. For each ancient Chinese character, the POS tag of it is obtained from the dictionary based on the POS tag of its aligned modern Chinese word, if it is paired with at least one modern Chinese word, we take the modern Chinese word with the highest alignment probability as its alignment object; if it is not paired with any modern Chinese word, we take it as a single character word and tag it as null value. After that, for adjacent ancient Chinese characters, we combine ones aligned with one modern Chinese word into one word. For example, in Figure \ref{fig-wa}, the characters ``裨'' (vice) and ``將'' (general) are both aligned with word 
\begin{CJK*}{UTF8}{gbsn}``副将''\end{CJK*} (vice-general), we combine them into one word ``裨將'' (vice-general); the character ``軍'' (army) is not aligned with any word, we take it as a single character word.

Although the word boundaries and POS tags can be projected form modern Chinese to ancient Chinese, there are still several issues to be solved: (1) Wrong word segmentation. For 
\begin{figure}[t]
  \vspace{0.2cm}
  \centering
  \centerline{\includegraphics[width=8.8cm]{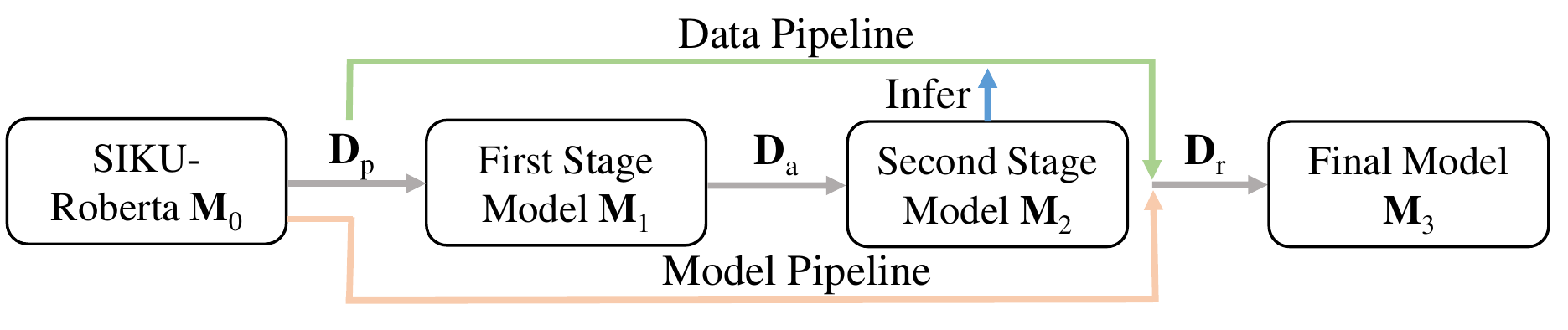}}
  \vspace{-0.3cm}
%
\caption{The pipeline of denoising and completing. `Infer' represents model's inference on the projected data $\rm \mathbf{D}_{p}$.}
\vspace{-2px}
\label{fig-pipe}
\end{figure}
example, in Figure \ref{fig-wa}, the word ``裨將軍'' (vice-general) is wrongly segmented as two words ``裨將'' (vice-general) and ``軍'' (army). (2) Wrong POS tagging results. The word ``與'' (with), whose POS tag in this sentence is `c' (conjunction), is wrongly tagged as `p' (preposition). (3) POS tag missing for some words, such as ``軍'' (army). To address these issues, we propose a framework of relabeling the projected data.

\subsection{Denoising and Completing by Relabeling}
\label{ssec:subhead2}

As shown in Figure \ref{fig-pipe}, the denoising and completing procedure can be divided into three stages. The initial model $\mathbf{M}_{0}$ is SIKU-Roberta\textsuperscript{\ref{2}}, which has been pretrained over a large unlabeled corpus of ancient Chinese collection, Siku-Quanshu.

\noindent$\bullet$~\textbf{First Stage: Training over Large Projected Data}. In first stage, we train the initial model $\mathbf{M}_{0}$ over the large projected data $\rm \mathbf{D}_{p}$ obtained from distant supervision to get first stage model $\mathbf{M}_{1}$. Due to the memorization effects of DNNs, which tend to fit majority (clean) patterns firstly and then overfit minority (noisy) patterns gradually \cite{arpit2017closer}\cite{bai2021understanding}, first stage model $\mathbf{M}_{1}$ has much knowledge of ancient Chinese and a little noise.

\noindent$\bullet$~\textbf{Second Stage: Training over Small Annotated Data}. In second stage, we use the small annotated data $\rm \mathbf{D}_{a}$ to decrease the influence of the noisy patterns learned by first stage model $\mathbf{M}_{1}$. After continuing to train first stage model $\mathbf{M}_{1}$ over the small data $\rm \mathbf{D}_{a}$, we get second stage model $\mathbf{M}_{2}$ having more knowledge and much lesser noise than first stage model $\mathbf{M}_{1}$. 
\end{CJK*}

\noindent$\bullet$~\textbf{Final Stage: Training over Large Relabeled Data}. In the process of training first stage model $\mathbf{M}_{1}$, words not tagged in distant supervision are not utilized in training and words wrongly tagged or segmented impair the training. To fully utilize the large projected data $\rm \mathbf{D}_{p}$, we use the second stage model $\mathbf{M}_{2}$ to relabel the large projected data $\rm \mathbf{D}_{p}$ and get the large relabeled data $\rm \mathbf{D}_{r}$. After relabeling, the noise of word boundaries and POS tags significantly decreases and the gaps in POS tags are filled. Finally, we retrain the initial model $\mathbf{M}_{0}$ over the large relabeled data $\rm \mathbf{D}_{r}$ to get final model $\mathbf{M}_{3}$. Our experiments show that final model $\mathbf{M}_{3}$ outperforms second stage model $\mathbf{M}_{2}$ at F1 scores of WSG and POS tagging.

\section{Experiments}
\label{sec:format0}

\subsection{Experimental Setup}
\label{ssec:subhead3}
\noindent$\bullet$~\textbf{Datasets:} The parallel corpus\footnote{https://github.com/NiuTrans/Classical-Modern} of modern Chinese and ancient Chinese contains 967257 pairs of sentences. Annotated datasets\textsuperscript{\ref{1}} contain three parts, one training dataset and two test datasets Test-A and Test-B \cite{li2022first}. The training dataset contains 8701 sentences sampled from \textit{Zuozhuan}. We divide the first nine tenths of the dataset into our training dataset $\rm \mathbf{D}_{a}$ and the last one tenth into our validation dataset. Test-A is also extracted from \textit{Zuozhuan}, containing 1594 sentences and designed to test model performance on the same book. Test-B is extracted from \textit{Shiji} and \textit{Zizhitongjian}, containing 2150 sentences and designed to test model performance on data, whose distribution deviates from the training dataset.


\noindent$\bullet$~\textbf{Comparison Methods:} We compare our designed method with other methods of data augmentation in EvaHan2022\textsuperscript{\ref{1}} evaluation campaign. Shen et al. \cite{shen2022data} use autoregressive language model to generate pseudo-data based on linearized labeled sentence \cite{ding2020daga} and ask the pretrained model to predict masked words to augment data. Tina et al. \cite{tian2022ancient} randomly mask the input words during training to implicitly augment data.

\subsection{Experimental Results}

\begin{table}[htbp]
\centering
\scalebox{0.82}
{
    \begin{tabular}
	{c|cc|cc}\hline
	
	\multirowcell{2}{Augmentation Method} & \multicolumn{2}{c|}{Test-A} & \multicolumn{2}{c}{Test-B} \\ 
	& WSG-F1 & POS-F1 & WSG-F1 & POS-F1 \\\hline
	
	Shen et al. \cite{shen2022data} & 94.81 & 89.87 & 88.42 & 79.53 \\ 
	Tian et al. \cite{tian2022ancient} & 94.73 & \textbf{90.93} & 89.19 & 83.48 \\ \hline

        $\rm \mathbf{D}_{p}$ ($\mathbf{M}_{1}$) & 90.22 & 75.78 & 88.22 & 74.75\\ 

	$\rm \mathbf{D}_{p}$+$\rm \mathbf{D}_{a}$ ($\mathbf{M}_{2}$) & 95.55 & 90.55 & 93.47 & 85.91 \\ 
	
	$\rm \mathbf{D}_{r}$ ($\mathbf{M}_{3}$) & \textbf{95.64} & 90.55 & \textbf{93.64} & \textbf{86.21}\\ \hline
    \end{tabular}
}
\caption{WSG and POS tagging results on Test-A and Test-B.}
\vspace{-5px}
\label{tab:main}
\end{table}

\noindent$\bullet$~\textbf{Main Results:} As shown in Table \ref{tab:main}, compared with Tina et al. \cite{tian2022ancient}, it's reasonable that our first stage model ($\mathbf{M}_{1}$) doesn't perform well on two test datasets because $\mathbf{M}_{1}$ is trained over weakly labeled data $\rm \mathbf{D}_{p}$. But on Test-A, our final model ($\mathbf{M}_{3}$) obtains an improvement of 0.91\% in WSG and only a decrease of 0.38\% in POS tagging; on Test-B, our final model ($\mathbf{M}_{3}$) obtains an improvement of 4.45\% in WSG and an improvement of 2.73\% in POS tagging. In general, our models $\mathbf{M}_{2}$ and $\mathbf{M}_{3}$ obtain improvements on both test datasets, especially on Test-B whose data distribution deviates from the training dataset, which proves the superiority of our method.

\begin{table}[htbp]
\centering
\scalebox{0.75}
{
	\begin{tabular}
	{c|c|cc|cc}\hline
	\multirowcell{2}{Data ratio}& \multirowcell{2}{Model} & \multicolumn{2}{c|}{Test-A} & \multicolumn{2}{c}{Test-B} \\ 
	& & WSG-F1 & POS-F1 & WSG-F1 & POS-F1 \\\hline
	
	\multirowcell{3}{$\rm R_{a}=25\%$\\$\rm R_{p}=100\%$} & $\rm \mathbf{D}_{a}$ ($\rm \mathbf{M}_{b}$) & 94.23 & 85.23 & 93.04 & 81.86 \\ 
	 & $\rm \mathbf{D}_{p}$+$\rm \mathbf{D}_{a}$ ($\mathbf{M}_{2}$) & 94.68 & 86.07 & 93.04 & 82.64 \\ 
	& $\rm \mathbf{D}_{r}$ ($\mathbf{M}_{3}$)& \textbf{94.91} & \textbf{86.57} & \textbf{93.28} & \textbf{83.09} \\ \hline
	
	\multirowcell{3}{$\rm R_{a}=50\%$\\$\rm R_{p}=100\%$} & $\rm \mathbf{D}_{a}$ ($\rm \mathbf{M}_{b}$) & 94.94 & 89.11 & \textbf{92.50} & 85.23 \\ 
	& $\rm \mathbf{D}_{p}$+$\rm \mathbf{D}_{a}$ ($\mathbf{M}_{2}$)& 94.94 & 88.98 & 92.28 & 85.08 \\ 
	& $\rm \mathbf{D}_{r}$ ($\mathbf{M}_{3}$) & \textbf{95.11} & \textbf{89.29} & 92.20 & \textbf{85.31}\\ \hline
	

	\multirowcell{3}{$\rm R_{a}=100\%$\\$\rm R_{p}=100\%$} & $\rm \mathbf{D}_{a}$ ($\rm \mathbf{M}_{b}$) & 95.62 & 90.24 & 93.25 & 85.21 \\ 
	& $\rm \mathbf{D}_{p}$+$\rm \mathbf{D}_{a}$ ($\mathbf{M}_{2}$) & 95.55 & 90.55 & 93.47 & 85.91 \\ 
	& $\rm \mathbf{D}_{r}$ ($\mathbf{M}_{3}$) & \textbf{95.64} & \textbf{90.55} & \textbf{93.64} & \textbf{86.21}\\ \hline
	\end{tabular}
}
\caption{The role of projected data in different data ratios. $\rm R_{a}$ represents the ration of annotated dataset we select. $\rm R_{p}$ represents the ration of projected dataset we select. $\rm \mathbf{M}_{b}$ is the baseline model only trained over samll annotated data $\rm \mathbf{D}_{a}$.}
\vspace{-5px}
\label{tab:Ann vs proj}
\end{table}

\begin{table}[htbp]
\centering
\scalebox{0.74}
{
	\begin{tabular}
	{c|c|cc|cc}\hline
	\multirowcell{2}{First stage\\training task}& \multirowcell{2}{Model} & \multicolumn{2}{c|}{Test-A} & \multicolumn{2}{c}{Test-B}\\
	& & WSG-F1 & POS-F1 & WSG-F1 & POS-F1 \\\hline
	
	\multirowcell{2}{WSG \& POS} 
	& $\rm \mathbf{D}_{p}$+$\rm \mathbf{D}_{a}$ ($\mathbf{M}_{2}$) & 95.55 & 90.55 & 93.47 & 85.91 \\ 
	& $\rm \mathbf{D}_{r}$ ($\mathbf{M}_{3}$) & \textbf{95.64} & \textbf{90.55} & 93.64 & \textbf{86.21} \\ \hline
	
	\multirowcell{2}{POS}
	& $\rm \mathbf{D}_{p}$+$\rm \mathbf{D}_{a}$ ($\mathbf{M}_{2}$) & 95.30 & 90.06 & 93.69 & 85.75 \\ 
	& $\rm \mathbf{D}_{r}$ ($\mathbf{M}_{3}$) & 95.36 & 90.11 & 93.89 & 86.15 \\ \hline
	
	\multirowcell{2}{WSG}
	& $\rm \mathbf{D}_{p}$+$\rm \mathbf{D}_{a}$ ($\mathbf{M}_{2}$) & 95.36 & 90.04 & 93.90 & 85.55 \\ 
	& $\rm \mathbf{D}_{r}$ ($\mathbf{M}_{3}$) & 95.38 & 89.71 & \textbf{94.06} & 85.28\\ \hline

	Null & $\rm \mathbf{D}_{a}$ ($\rm \mathbf{M}_{b}$) & 95.62 & 90.24 & 93.25 & 85.21 \\ \hline

	\end{tabular}
}
\caption{Experiments on multitask data augmentation.}
\vspace{-10px}
\label{tab-multi}
\end{table}
\noindent$\bullet$~\textbf{Analysis on Different Data Ratios:} As shown in Table \ref{tab:Ann vs proj}, we set the ratio of projected data to 100\% and gradually increase the ratio of annotated data, and the F1 scores gradually increase accordingly. Compared with $\rm \mathbf{M}_{b}$, it is obvious that $\rm \mathbf{M}_{3}$ obtains improvements (1.34\% and 1.23\%) of POS tagging in {\{$\rm R_{a}=25\%$\}}, higher than that (0.31\% and 1.00\%) in {\{$\rm R_{a}=100\%$\}}, which shows that our method performs well in low resource scenario. In almost all settings, the ranking of model is $\rm \mathbf{M}_{3}$ > $\rm \mathbf{M}_{2}$ > $\rm \mathbf{M}_{b}$, so the ranking of knowledge provided by data is $\rm \mathbf{D}_{r}$ > $\rm \mathbf{D}_{p}$+$\rm \mathbf{D}_{a}$ > $\rm \mathbf{D}_{a}$, which proves the effectiveness of distant supervision and relabeling the large projected data $\rm \mathbf{D}_{p}$.

\noindent$\bullet$~\textbf{Analysis on Augmentation of WSG and POS Tagging:} In distant supervision, we project the word boundaries and POS tags of modern Chinese to ancient Chinese using word alignment. To further explore the role of WSG and POS tagging of projected data in data augmentation, we perform experiments on different combinations of training tasks in the first stage training under data ratio {\{$\rm R_{a}=100\%$, $\rm R_{p}=100\%$\}}. As shown in Table \ref{tab-multi}, training under WSG and POS tagging occupies the first place, training under either WSG or POS tagging comes second, which indicates the effectiveness of the projection of word boundaries and POS tags. 

\section{CONCLUSION}

In this paper, we propose a method of augmentation of ancient Chinese WSG and POS tagging data using distant supervision. Besides, we use the method of relabeling to reduce the noise introduced by distant supervision. Experiments show the effectiveness of distant supervision and relabeling.

\noindent$\bullet$~\textbf{Acknowledgements:} This research is supported by the National Natural Science Foundation of China (No.62106105), the CCF-Tencent Open Research Fund (No.RAGR20220122), the Scientific Research Starting Foundation of Nanjing University of Aeronautics and Astronautics (No.YQR21022), and the High Performance Computing Platform of Nanjing University of Aeronautics and Astronautics.

\bibliographystyle{IEEEbib}
\bibliography{strings,refs}

\begin{thebibliography}{10}

\bibitem{li2022first}
Bin Li, Yiguo Yuan, Jingya Lu, Minxuan Feng, Chao Xu, Weiguang Qu, and Dongbo
  Wang,
\newblock ``The first international ancient chinese word segmentation and pos
  tagging bakeoff: Overview of the evahan 2022 evaluation campaign,''
\newblock in {\em Proceedings of the Second Workshop on Language Technologies
  for Historical and Ancient Languages}, 2022, pp. 135--140.

\bibitem{ding2020daga}
Bosheng Ding, Linlin Liu, Lidong Bing, Canasai Kruengkrai, Thien~Hai Nguyen,
  Shafiq Joty, Luo Si, and Chunyan Miao,
\newblock ``Daga: Data augmentation with a generation approach for low-resource
  tagging tasks,''
\newblock {\em arXiv preprint arXiv:2011.01549}, 2020.

\bibitem{shen2022data}
Yutong Shen, Jiahuan Li, Shujian Huang, Yi~Zhou, Xiaopeng Xie, and Qinxin Zhao,
\newblock ``Data augmentation for low-resource word segmentation and pos
  tagging of ancient chinese texts,''
\newblock {\em LT4HALA 2022}, p. 169, 2022.

\bibitem{tian2022ancient}
Yanzhi Tian and Yuhang Guo,
\newblock ``Ancient chinese word segmentation and part-of-speech tagging using
  data augmentation,''
\newblock in {\em Proceedings of the Second Workshop on Language Technologies
  for Historical and Ancient Languages}, 2022, pp. 146--149.

\bibitem{mintz2009distant}
Mike Mintz, Steven Bills, Rion Snow, and Dan Jurafsky,
\newblock ``Distant supervision for relation extraction without labeled data,''
\newblock in {\em Proceedings of the Joint Conference of the 47th Annual
  Meeting of the ACL and the 4th International Joint Conference on Natural
  Language Processing of the AFNLP}, 2009, pp. 1003--1011.

\bibitem{che2019word}
Chao Che, Hanyu Zhao, Xiaoting Wu, Dongsheng Zhou, and Qiang Zhang,
\newblock ``A word segmentation method of ancient chinese based on word
  alignment,''
\newblock in {\em CCF International Conference on Natural Language Processing
  and Chinese Computing}. Springer, 2019, pp. 761--772.

\bibitem{yarowsky2001inducing}
David Yarowsky and Grace Ngai,
\newblock ``Inducing multilingual pos taggers and np bracketers via robust
  projection across aligned corpora,''
\newblock in {\em Second Meeting of the North American Chapter of the
  Association for Computational Linguistics}, 2001.

\bibitem{fang2016learning}
Meng Fang and Trevor Cohn,
\newblock ``Learning when to trust distant supervision: An application to
  low-resource pos tagging using cross-lingual projection,''
\newblock {\em arXiv preprint arXiv:1607.01133}, 2016.

\bibitem{wang2014cross}
Mengqiu Wang and Christopher~D Manning,
\newblock ``Cross-lingual projected expectation regularization for weakly
  supervised learning,''
\newblock {\em Transactions of the Association for Computational Linguistics},
  vol. 2, pp. 55--66, 2014.

\bibitem{mcdonald2013universal}
Ryan McDonald, Joakim Nivre, Yvonne Quirmbach-Brundage, Yoav Goldberg, Dipanjan
  Das, Kuzman Ganchev, Keith Hall, Slav Petrov, Hao Zhang, Oscar
  T{\"a}ckstr{\"o}m, et~al.,
\newblock ``Universal dependency annotation for multilingual parsing,''
\newblock in {\em Proceedings of the 51st Annual Meeting of the Association for
  Computational Linguistics (Volume 2: Short Papers)}, 2013, pp. 92--97.

\bibitem{das2011unsupervised}
Dipanjan Das and Slav Petrov,
\newblock ``Unsupervised part-of-speech tagging with bilingual graph-based
  projections,''
\newblock 2011.

\bibitem{tackstrom2013token}
Oscar T{\"a}ckstr{\"o}m, Dipanjan Das, Slav Petrov, Ryan McDonald, and Joakim
  Nivre,
\newblock ``Token and type constraints for cross-lingual part-of-speech
  tagging,''
\newblock {\em Transactions of the Association for Computational Linguistics},
  vol. 1, pp. 1--12, 2013.

\bibitem{arpit2017closer}
Devansh Arpit, Stanis{\l}aw Jastrz{\k{e}}bski, Nicolas Ballas, David Krueger,
  Emmanuel Bengio, Maxinder~S Kanwal, Tegan Maharaj, Asja Fischer, Aaron
  Courville, Yoshua Bengio, et~al.,
\newblock ``A closer look at memorization in deep networks,''
\newblock in {\em International conference on machine learning}. PMLR, 2017,
  pp. 233--242.

\bibitem{bai2021understanding}
Yingbin Bai, Erkun Yang, Bo~Han, Yanhua Yang, Jiatong Li, Yinian Mao, Gang Niu,
  and Tongliang Liu,
\newblock ``Understanding and improving early stopping for learning with noisy
  labels,''
\newblock {\em Advances in Neural Information Processing Systems}, vol. 34, pp.
  24392--24403, 2021.

\bibitem{shi2010crf}
Min Shi, Bin Li, and Xiaohe Chen,
\newblock ``Crf based research on a unified approach to word segmentation and
  pos tagging for pre-qin chinese,''
\newblock {\em Journal of Chinese Information Processing}, vol. 2, no. 24, pp.
  39--45, 2010.

\bibitem{lafferty2001conditional}
John Lafferty, Andrew McCallum, and Fernando~CN Pereira,
\newblock ``Conditional random fields: Probabilistic models for segmenting and
  labeling sequence data,''
\newblock 2001.

\bibitem{forney1973viterbi}
G~David Forney,
\newblock ``The viterbi algorithm,''
\newblock {\em Proceedings of the IEEE}, vol. 61, no. 3, pp. 268--278, 1973.

\bibitem{che2010ltp}
Wanxiang Che, Zhenghua Li, and Ting Liu,
\newblock ``Ltp: A chinese language technology platform,''
\newblock in {\em Coling 2010: Demonstrations}, 2010, pp. 13--16.

\bibitem{della1994mathematics}
Vincent~J Della~Pietra,
\newblock ``The mathematics of statistical machine translation: Parameter
  estimation,''
\newblock {\em Using Large Corpora}, p. 223, 1994.

\bibitem{jinteng2013construction}
Liu Jinteng, Song Yan, and Xia Fei,
\newblock ``The construction of a segmented and part-of-speech tagged archaic
  chinese corpus: A case study on huainanzi,''
\newblock {\em Journal of Chinese Information Processing}, vol. 27, no. 6, pp.
  6--15, 2013.

\end{thebibliography}
\end{document}